# An Online Boosting Algorithm with Theoretical Justifications

Shang-Tse Chen[†]　　　　　　　　　　　　　　　　　　　　　　　　　b95100@csie.ntu.edu.tw
Hsuan-Tien Lin[‡]　　　　　　　　　　　　　　　　　　　　　　　　　　htlin@csie.ntu.edu.tw
Chi-Jen Lu[†]　　　　　　　　　　　　　　　　　　　　　　　　　　　cjlu@iis.sinica.edu.tw

[†] Institute of Information Science, Academia Sinica, Taipei, Taiwan
[‡] Department of Computer Science and Information Engineering, National Taiwan University

## Abstract

We study the task of online boosting — combining online weak learners into an online strong learner. While batch boosting has a sound theoretical foundation, online boosting deserves more study from the theoretical perspective. In this paper, we carefully compare the differences between online and batch boosting, and propose a novel and reasonable assumption for the online weak learner. Based on the assumption, we design an online boosting algorithm with a strong theoretical guarantee by adapting from the offline SmoothBoost algorithm that matches the assumption closely. We further tackle the task of deciding the number of weak learners using established theoretical results for online convex programming and predicting with expert advice. Experiments on real-world data sets demonstrate that the proposed algorithm compares favorably with existing online boosting algorithms.

## 1. Introduction

Boosting is one of the most powerful and popular ensemble-learning techniques in the setting of batch learning. The technique is an important topic from both the theoretical and practical perspective. On the theoretical side, boosting identifies the least (weakest) assumption on the learner to make learning possible (Freund & Schapire, 1996; Schapire et al., 1998; Mukherjee et al., 2011), and the assumption can be used to facilitate the analysis of existing algorithms and the design of new ones. On the practical side, boosting allows re-using of existing (weak) learning algorithms in an efficient manner to improve performance, which matches the needs of many real-world applications (Schapire & Singer, 2000; Kudo et al., 2004; He & Thiesson, 2007).

Online learning, as opposed to batch learning, is another important topic in machine learning. Online learning does not require a fixed set of training data on hand but processes streaming examples one by one, which also fits the needs of many real-world applications. For example, a spam filtering system might need to continuously adjust its filtering rules based on the ever-changing spam tactics. Online learning also has its advantages in handling large-scale data sets, since it does not need to load the whole data set into the memory. The success of boosting in batch learning and the many merits of online learning inspire the study of online boosting — a combination of the two. For instance, consider an online spam classifier that has been working reasonably well, can we "boost" the performance by combining a couple of those classifiers?

The work of Oza & Russell (2001) is one of the first to use boosting in the online setting, and it was argued that the given algorithm under some condition could converge to the popular Adaptive Boosting approach (AdaBoost; Freund & Schapire, 1997) as the number of weak learners and training examples approaches infinity. Online boosting also achieved great success in many real-world applications, especially in the field of computer vision (Grabner & Bischof, 2006), due to its simplicity and efficiency. Many other online boosting algorithms have been proposed to tackle different application needs, such as semi-supervised learning (Grabner et al., 2008), multi-instance learning (Babenko et al., 2009b), and feature selection (Liu & Yu, 2007).

Nevertheless, relatively few existing studies discuss the theoretical behaviors of online boosting algorithms, as opposed to their offline counterparts. While many works on online boosting try to approximate AdaBoost or other batch boosting algorithms as closely as possible, they ignore the intrinsic differences between online learning and batch learning. In this paper, we carefully compare these differences, which in turn leads to different design strategies of the algorithms.

We start by re-examining the foundation of boosting algorithms — the weak learning assumption, which says that under any distribution of the data, the weak learner can perform better than random guessing. While this is a slightly strong but reasonable assumption in the batch setting, it is far from realistic in the online setting because the online weak learners are more restricted regarding the information available. We thus propose a new and more reasonable assumption that requires the online weak learners to perform well only with respect to "smoother" distributions. Based on this new assumption, we try to find a boosting algorithm that assigns example weights in a more "conservative" and online manner. One particular boosting algorithm that not only fits our requirements with slight modifications but also comes with simple and elegant theoretical analysis is *SmoothBoost* (Servedio, 2003). In this paper, we extend it to an online boosting algorithm.

Another difficulty of online boosting is that we have to determine beforehand the number of weak learners we would like to combine. The danger is that if we include too many weak learners, the outcome may be dominated by the poor predictions made by the many redundant weak learners which do not learn well. We mitigate this problem by giving different voting weights to different weak learners and we determine these weights dynamically using *Online Convex Programming* (Zinkevich, 2003) and the framework of *Predicting with Expert Advice* (Cesa-Bianchi & Lugosi, 2006), both of which are well-established techniques in online learning.

Our final online boosting algorithms have the nice feature that theoretical guarantees can actually be shown, just as in the batch setting. In particular, we show that given online weak learners which can predict slightly better than random guessing with respect to "smooth" distributions, our online boosting algorithms can combine them to achieve a small error rate. In addition, we also perform experiments on several benchmark data sets, and the results show that our algorithms not only are theoretically well-founded, but also work well empirically on these real-world data sets.

## 2. Online versus Batch Boosting

We consider the online learning problem in which an online learner must process a stream of examples $(x_1, y_1), \ldots, (x_T, y_T) \in \mathbb{R}^d \times \{-1, 1\}$ in the following way. In step $t$, the online learner receives $x_t$ and is required to predict its label, and after the prediction the true label $y_t$ is revealed. We study the possibility of designing such an online learner using the boosting approach. That is, if we have online weak learners which can make predictions slightly better than random guessing, can we combine them to obtain an online strong learner which can make correct predictions for all but a small fraction of the examples?

Before formally describing our online boosting framework, let us first recall that of batch boosting. In the batch setting, the whole set $S$ of labeled examples is available at the beginning, and the boosting algorithm proceeds for some $N$ rounds as follows. In round $i$, it chooses a distribution $p^{(i)}$ over $S$ and gives $S$ and $p^{(i)}$ to a batch weak learner, which has the whole $S$ and $p^{(i)}$ available and produces a weak hypothesis $h^{(i)}$. After the $N$ rounds, it combines the $N$ weak hypotheses to produce the final strong hypothesis, which takes the form of $H(x) = \text{sign}(\sum_{i=1}^{N} \alpha^{(i)} h^{(i)}(x))$, where $\alpha^{(i)} \in \mathbb{R}$ is the voting weight of $h^{(i)}$. There are boosting algorithms which can achieve a small error rate, defined as $\frac{1}{T}|\{t : H(x_t) \neq y_t\}|$, if each weak hypothesis $h^{(i)}$ has a positive advantage, defined as $\sum_{t=1}^{T} p_t^{(i)} y_t h^{(i)}(x_t)$.

Now, in our online boosting framework, the examples of $S$ only become available one at a time, and the boosting algorithm as well as the weak learners must work in an online fashion. Thus, the boosting algorithm cannot call the weak learner sequentially in $N$ rounds as in the batch setting and must run $N$ weak learners in parallel. That is, for each received example $(x_t, y_t)$, the boosting algorithm must update the $N$ weak learners right away before seeing the remaining examples. To do that, one would like the boosting algorithm to send a measure $p_t^{(i)}$ of $(x_t, y_t)$ to the $i$-th weak learner. However, it does not seem easy to determine a good "measure" of an example without seeing the remaining examples, and a somewhat easier but sufficient task is to send a "weight" $w_t^{(i)}$ of $(x_t, y_t)$, so that $w_t^{(i)} / \sum_{t=1}^{T} w_t^{(i)}$ corresponds to the measure $p_t^{(i)}$ of $(x_t, y_t)$ for the $i$-th weak learner. Then, after the update, the $i$-th weak learner returns a weak hypothesis $h_{t+1}^{(i)}$, and the boosting algorithm predicts the next example $x_{t+1}$ by $H_{t+1}(x_{t+1}) = \text{sign}(\sum_{i=1}^{N} \alpha_{t+1}^{(i)} h_{t+1}^{(i)}(x_{t+1}))$, where $\alpha_{t+1}^{(i)} \in \mathbb{R}$ is the voting weight of $h_{t+1}^{(i)}$. As in the batch setting, we would like to have an online boosting algorithm which can achieve a small error rate, defined as $\frac{1}{T}|\{t : H_t(x_t) \neq y_t\}|$, if each weak learner has some positive advantage, defined as $\sum_{t=1}^{T} p_t^{(i)} y_t h_t^{(i)}(x_t)$.

However, there appear to be some difficulties in designing such an online boosting algorithm. First, for each example, its weight for each weak learner must be determined before seeing the remaining examples. This rules out the use of the weighting schemes of some batch boosting algorithms. Next, the weights must satisfy some additional property in order to expect an online weak learner to have a positive advantage. To see this, if we take the extreme case that the first example has weight 1 and all others have weight 0, it is unrealistic to expect an online weak learner to have a positive advantage. Note that this would not be a problem in the batch setting, since the weak learner can read all the examples as well as their labels before coming up with a hypothesis. Finally, it is not clear how the online boosting algorithm can choose the appropriate number $N$ of weak learners. We only know an upper bound for $N$ which may be much larger than the appropriate one, but if we combine too many weak learners, the result may be dominated by the poor predictions of the many weak learners which should not be included. This is not a problem in the batch setting, as the boosting algorithm proceeds in rounds, including a new hypothesis in each round, and stops once the new weak hypothesis fails to have the required advantage.

For simplicity, let us assume that each $x_t$ lies within the unit $L_2$-ball, so that $\|x_t\|_2 \leq 1$, and each weak hypothesis $h_t^{(i)}$ comes from some set $\mathcal{H}$ of functions mapping from the unit $L_2$-ball to the interval $[-1, 1]$. We will use the notation $[T]$ to denote the set $\{1, \ldots, T\}$ for a positive integer $T$.

## 3. Online Weak Learners

In this section, we address the second difficulty discussed in the previous section and study the condition for an online weak learner to have a positive advantage. Let us consider the case that $\mathcal{H}$, the weak hypotheses space, consists of linear functions, so that each $h \in \mathcal{H}$ can be seen as a vector in $\mathbb{R}^d$, with $h(x)$ defined as $\langle h, x \rangle$, the inner product of the vectors $h$ and $x$. For simplicity, let us assume $\|h\|_2 \leq 1$ for every $h \in \mathcal{H}$.

We can reduce the problem of finding a good online weak learner to the well-known online linear optimization problem as follows. With the $T$ examples of $S$ arriving sequentially, the weak learner in round $t$ is given the data $x_t$ as well as its weight $w_t$, and it then produces a hypothesis $h_t \in \mathcal{H}$ and after that receives a reward $r_t(h_t) = w_t y_t h_t(x_t) = w_t y_t \langle h_t, x_t \rangle$. Note that the reward function $r_t$ is linear in $h_t$. Therefore, we can apply the gradient descent algorithm of (Zinkevich, 2003) to produce $h_t$ in round $t$, and a standard regret analysis shows that for some constant $c > 0$,

$$\sum_{t=1}^T w_t y_t h_t(x_t) \geq \sum_{t=1}^T w_t y_t h(x_t) - \sqrt{c \sum_{t=1}^T w_t^2}, \quad (1)$$

for any fixed hypothesis $h \in \mathcal{H}$ chosen by an offline algorithm. Let $|w| = \sum_{t=1}^T w_t$, and by dividing both sides above by $|w|$, we have

$$\sum_{t=1}^T p_t y_t h_t(x_t) \geq \sum_{t=1}^T p_t y_t h(x_t) - \sqrt{c \sum_{t=1}^T w_t^2/|w|^2},$$

where $p_t = w_t/|w|$ is the measure of $(x_t, y_t)$. The term $\sum_{t=1}^T p_t y_t h_t(x_t)$ is the advantage of the online weak learner. The term $\sum_{t=1}^T p_t y_t h(x_t)$ is the advantage of the offline hypothesis $h$, and let us assume for now that this advantage is at least $3\gamma > 0$. Moreover, suppose the weights are large in the sense that they satisfy the following condition:

$$|w| \geq c/\gamma^2 \text{ and } w_t \in [0, 1] \text{ for every } t, \quad (2)$$

where $c$ is the constant in (1). Then the advantage of the online weak learner becomes

$$\sum_{t=1}^T p_t y_t h_t(x_t) \geq 3\gamma - \sqrt{c|w|/|w|^2} \geq 2\gamma, \quad (3)$$

where in the inequality we use $\sum_{t=1}^T w_t^2 \leq |w|$ as $w_t \in [0, 1]$. Note that in addition to being sufficient, one can also show that the condition (2) is necessary to guarantee (3), using standard approaches for proving regret lower bounds. This motivates us to introduce the following assumption.

**Assumption 1.** *There exists an online weak learner which can achieve advantage $2\gamma > 0$ for any sequence of examples and weights satisfying the condition (2).*

Based on the discussion above, we have the following.

**Lemma 1.** *Suppose for any sequence of examples and weights satisfying the condition (2), there exists an offline linear hypothesis with advantage $3\gamma > 0$. Then Assumption 1 holds.*

Note that large weights satisfying (2) give rise to distributions which are "smooth" in the sense that each example has measure at most $1/|w| \leq \gamma^2/c$. This excludes the extreme case discussed in the previous section and makes possible for an online weak learner to have a positive advantage. The concept of smoothness has also been applied to boosting in several frameworks, such as noise-tolerant learning (Servedio, 2003) and agnostic learning (Feldman, 2010), but to the best of our knowledge, this is the first work that incorporates this idea into the problem of online boosting.

## 4. Our Online Boosting Algorithm

In this section, we show how to choose weights for examples and how to combine hypotheses from weak learners in order to obtain an online boosting algorithm. Recall the framework of online boosting described in Section 2. Suppose Assumption 1 holds and let WL be such an online weak learner with advantage $2\gamma$. We will run $N$ copies of WL as our $N$ weak learners, for some $N$ to be determined later.

First, we would like to produce weights satisfying the condition (2). It is known that AdaBoost does not always produce such weights (Bshouty et al., 2002). One may try to scale up or down the weights to satisfy the condition, but the scaling factors often can only be determined after seeing all the examples, which does not work in the online setting. Fortunately, we can adopt the weighting scheme similar to SmoothBoost (Servedio, 2003) by choosing

$$w_t^{(i+1)} = \min\left\{(1-\gamma)^{z_t^{(i)}/2}, 1\right\}, \quad (4)$$

where $z_t^{(i)} = z_t^{(i-1)} + y_t h_t^{(i)}(x_t) - \theta$, with $\theta = \gamma/(2+\gamma)$, and $z_t^{(0)} = 0$ for $t \in [T]$. It is easy to verify that given $(x_t, y_t)$ at step $t$, one can compute the weights $w_t^{(1)}, \ldots, w_t^{(N)}$. Moreover, the following lemma shows that we only need hypotheses from those weak learners associated with large weights.

**Lemma 2.** *For any $i \in [N]$ and $t \in [T]$, define*

$$f_t^{(i)}(x) = \frac{1}{i}\sum_{j=1}^{i} h_t^{(j)}(x) \text{ and } H_t^{(i)}(x) = \text{sign}\left(f_t^{(i)}(x)\right).$$

*Let $\delta \in [0,1]$ and let $k$ be the largest number such that $|w^{(i)}| \geq \delta T$ for every $i < k$. Then*

$$\frac{1}{T}\left|\{t : H_t^{(k)}(x_t) \neq y_t\}\right| \leq \frac{1}{T}\left|\{t : y_t f_t^{(k)}(x_t) \leq \theta\}\right| < \delta.$$

The idea is that the algorithm assigns large weights to those incorrectly predicted examples, so there can not be many of them if the sum of weights is small. Note that when $T \geq c/(\delta\gamma^2)$, for the constant $c$ in (2), the weights for the $i$-th weak learner, for $i < k$, are then large enough to satisfy the condition (2) since $|w^{(i)}| \geq \delta T \geq c/\gamma^2$. The next lemma gives an upper bound for the parameter $k$.

**Lemma 3.** *Suppose Assumption 1 holds and $T \geq c/(\delta\gamma^2)$. Then the parameter $k$ in Lemma 2 is at most $\mathcal{O}(1/(\delta\gamma^2))$.*

We omit the proofs of these two lemmas because they are almost identical to those for Theorem 2 and Theorem 3 in Servedio (2003) respectively.

Lemma 3 provides us an upper bound on the number $N$ of weak learners. However, the problem in the online setting is that we have to determine $N$ beforehand, and empirically we see that setting $N$ to this upper bound and using the function $H^{(N)}$ in Lemma 2 for prediction would not get very good performance. The reason is that if $N$ exceeds the actual number $k$ in Lemma 3, we could include too many weak learners which receive examples with small weights and would not update much. These weak learners might not learn well enough and taking them into the ensemble would therefore hinder the performance. Next, we describe two approaches to solve this problem.

The first is to give different voting weights for different weak learners, intuitively with larger weights to better weak learners. We set $N$ to the upper bound given in Lemma 3, and at step $t$, we find some voting weight $\alpha_t^{(i)}$ for the $i$-th weak learner, and predict $x_t$ by

$$H_t(x_t) = \text{sign}\left(\sum_{i=1}^{N} \alpha_t^{(i)} h_t^{(i)}(x)\right).$$

To find appropriate $\alpha_t = (\alpha_t^{(1)}, \ldots, \alpha_t^{(N)})$, we reduce the task to the *Online Convex Programming* (OCP) problem, using the $N$-dimensional probability simplex as the feasible set and defining the loss function at step $t$ as

$$\ell_t(\alpha) = \max\left\{0, \theta - \sum_{i=1}^{N} \alpha^{(i)} y_t h_t^{(i)}(x_t)\right\},$$

which is a convex function of $\alpha = (\alpha^{(1)}, \ldots, \alpha^{(N)})$. From Lemma 2, we know the existence of an $\bar{\alpha} = (\bar{\alpha}^{(1)}, \ldots, \bar{\alpha}^{(N)})$, with $\bar{\alpha}^{(i)} = 1/k$ for $i < k$ and $\bar{\alpha}^{(i)} = 0$ for $i \geq k$, such that

$$\sum_{t=1}^{T} \ell_t(\bar{\alpha}) < (\theta+1)\delta T.$$

Thus, if we use the gradient descent algorithm of (Zinkevich, 2003) to produce $\alpha_t$ at step $t$, we can have

$$\sum_{t=1}^{T} \ell_t(\alpha_t) < (\theta+1)\delta T + \sqrt{cT}.$$

To relate this bound with the error rate of $H_t$, note that if $H_t(x_t) \neq y_t$, then $\ell_t(\alpha_t) \geq \theta$, which implies $|\{t : H_t(x_t) \neq y_t\}| \leq \frac{1}{\theta}\sum_{t=1}^{T} \ell_t(\alpha_t)$. Thus, we have

$$\frac{1}{T}|\{t : H_t(x_t) \neq y_t\}| < \frac{\theta+1}{\theta}\delta + \frac{\sqrt{c}}{\theta\sqrt{T}}.$$

The complete algorithm is given in Algorithm 1.

**Algorithm 1** Online Boosting with OCP
>  **Input:** streaming examples $(x_1, y_1), \ldots, (x_T, y_T)$
>  parameters $0 < \delta < 1$, $0 \leq \theta < \gamma < \frac{1}{2}$
>  online weak learner WL
>  **Initialize:** $z_t^{(0)} = 0$ for $t \in [T]$
>  $w_t^{(1)} = 1$ for $t \in [T]$
>  $\alpha_1^{(i)} = \frac{1}{N}$ for $i \in [N]$
>  select random $h_1^{(i)} \in \mathcal{H}$ for $i \in [N]$
>  **for** $t = 1$ **to** $T$ **do**
>  Define $f_t(x) = \sum_{i=1}^{N} \alpha_t^{(i)} h_t^{(i)}(x)$
>  Predict with $H_t(x) = \text{sign}(f_t(x))$
>  **if** $y_t f_t(x_t) < \theta$ **then**
>  $\alpha_{t+1}^{(i)} = \alpha_t^{(i)} + \eta_t y_t h_t^{(i)}(x_t)$, for $i \in [N]$
>  project $\alpha_{t+1}$ back into probability simplex
>  **end if**
>  **for** $i = 1$ **to** $N$ **do**
>  $h_{t+1}^{(i)} \leftarrow \text{WL}\left(h_t^{(i)}, (x_t, y_t), w_t^{(i)}\right)$
>  $z_t^{(i)} = z_t^{(i-1)} + y_t h_t^{(i)}(x_t) - \theta$
>  $w_t^{(i+1)} = \min\left\{(1-\gamma)^{z_t^{(i)}/2}, 1\right\}$
>  **end for**
>  **end for**

The second approach to combine weak learners is to use the framework of *Predicting with Expert Advice*. However, if we simply use each weak learner as an expert, we can only perform comparably to the best weak learner, which is inadequate for our goal of competing with the best combination of weak learners. So we instead construct another $N$ experts, with the $i$-th expert using the function $H_t^{(i)}$ in Lemma 2 to predict $x_t$. Then, by running the weighted majority algorithm (Littlestone & Warmuth, 1994; Freund & Schapire, 1997) on these $N$ experts, the expected regret with respect to the best expert is at most $2\sqrt{T \ln N}$. Since the best expert according to Lemma 3 has error rate at most $\delta$, the expected error rate of our algorithm is thus at most $\delta + 2\sqrt{(\ln N)/T}$. The resulting boosting algorithm based on this approach can be easily modified from Algorithm 1.

We summarize our result as follows.

**Theorem 1.** *Suppose Assumption 1 holds and $T \geq c/(\delta\gamma^2)$ for a large enough constant $c$. Then there is an online boosting algorithm which uses $\mathcal{O}(1/(\delta\gamma^2))$ copies of weak learners and achieves an error rate of $\mathcal{O}(\delta)$.*

Theorem 1 works for any set of weak hypotheses, assuming the existence of online weak learners which can predict with such hypotheses with a positive advantage when given examples with large enough weights (Assumption 1). When specialized to linear weak hypotheses, we can lift the assumption of having an algorithm for finding good hypotheses, and replace it with the assumption of the mere existence of good hypotheses. More precisely, using Lemma 1 together with Theorem 1, we have the following.

**Corollary 1.** *Suppose for any sequence of examples and weights satisfying the condition (2), there is an offline linear hypothesis with advantage $3\gamma$, and $T \geq c/(\delta\gamma^2)$ for a large enough constant $c$. Then there is an online algorithm which achieves an error rate of $\mathcal{O}(\delta)$.*

## 5. Experiments

In this section, we compare the empirical performance of the proposed algorithms with two other leading ones in online boosting. We do not compare with some other leading algorithms because of the different settings between those and our proposed ones. For instance, Grbovic & Vucetic (2011) proposed the incremental boosting algorithm, which is allowed to store previous examples while our proposed algorithms only rely on the newest arriving example. Another case is the algorithm in Pelossof et al. (2008), which assumes the weak learners to be static, i.e. pre-trained offline, and only updates the weights for combining the learners, while our proposed algorithms allows online, dynamic weak learners. Yet another family of online boosting algorithms are proposed in Babenko et al. (2009a), which requires the weak learners to be updated using stochastic gradient descent so that the weak learners can be chained with optimizing a choice of loss function, while our proposed algorithms treat weak learners in a black-box manner with minimum assumptions. Below we briefly describe the two algorithms we compare with.

### 5.1. Compared Algorithms

1. **Online AdaBoost** (Oza & Russell, 2001) uses a Poisson sampling process to approximate the weighting method of AdaBoost. It guarantees that by using *lossless* online weak learner, which outputs the same hypothesis when trained online and offline for the same training set, their algorithm will converge to AdaBoost as the number of models and training examples approaches infinity. Nevertheless, the algorithm only ensures a good hypothesis asymptotically, but is not proved to achieve low regret *during* online learning. We will denote the algorithm as *OzaBoost*.

2. **Online GradientBoost** (Leistner et al., 2009) is an online variant of GradientBoost (Friedman, 2000), which uses functional gradient descent to

An Online Boosting Algorithm with Theoretical Justificationsproperly.

decide the optimal example weights and greedily minimizes the loss function of interest. One special trick in the algorithm is to use some "selectors." Each selector builds on top of a set of $K$ weak learners to choose the best learner for the ensemble at each online learning iteration. The idea of using selectors is mainly for doing feature selection (Grabner & Bischof, 2006) along the online learning process, and only fits the setting of our proposed algorithm when $K = 1$. We take such a $K$ for a fair comparison, and run the algorithm with the *logit* loss function that has consistently been the best choice in existing experimental studies (Leistner et al., 2009). We will denote the algorithm as *OGBoost*.

### 5.2. Weak Learners

We choose two different weak learners in our experiments in order to take a closer look at the boosting ability of our proposed algorithms. The first one is *Perceptron* (Rosenblatt, 1962), a standard and famous online learning algorithm which, like our analysis in Section 3, takes a hypothesis set of linear functions. The second weak learner we choose is *Naive Bayes*, which is a lossless online algorithm, a crucial condition for ensuring the convergence of OzaBoost.

### 5.3. Results

We tested our algorithms using 12 binary classification benchmark data sets (Frank & Asuncion, 2010), downloaded in the processed format from the LIBSVM repository.[1] For each online boosting algorithm, we couple it with 100 weak learners and record its error in an online setting for 5 trials, where each trial consists of a different random ordering of the examples. We then report the average error over the 5 trials. We have tried some different values of $\gamma$ and have confirmed that the performance of the proposed algorithms is quite stable to the choice of $\gamma$. Therefore, for simplicity we only show the results with $\gamma = 0.1$.

We first show that our algorithm can really boost the performance of the two different weak learners and outperform the other online boosting algorithms for most data sets, which are summarized in Tables 1 and 2. The proposed algorithm, when using only the uniform weighting scheme, is denoted as OSBoost (Online Smooth-Boost). Other variants will be examined later in this section. The bold entries in the OSBoost column indicates that OSBoost improves the performance over a single weak learner, and a '*' in the col-umn indicates the best performing boosting algorithm.

For the *Perceptron* weak learner in Table 1, our proposed OSBoost is consistently better than a single weak learner across all the data sets. Furthermore, on 8 out of the 12 data sets, OSBoost is better than the other two leading algorithms in online boosting. The performance difference is especially evident on large data sets. For the *Naive Bayes* weak learner in Table 2, our proposed OSBoost remains to be the best choice: boosting the performance of a single weak learner and continuing to be superior to the other two algorithms with a big difference in performance.

We then discuss the effectiveness of OSBoost.OCP and OSBoost.EXP, the two variants of our algorithm using *Online Convex Programming* and *Predicting with Expert Advice* as described in Section 4, respectively. The results are shown in Table 3 and 4. The bold entries in the OCP and EXP columns indicate that the variant improves the performance of the basic, uniformly-weighted OSBoost. First of all, we see that the basic OSBoost readily performs quite well, while the variants can sometimes result in a marginal gain of performance. The improvements of OSBoost.EXP over the basic OSBoost is more prominent when the data set is small, which is because OSBoost.EXP can be implicitly adapted to use fewer weak learners in a randomized manner to avoid overfitting the small set. The improvements of OSBoost.OCP over the basic OSBoost, on the other hand, happen mostly on larger data sets, which is in accordance to our analysis since the average error would diminish in *Online Convex Programming* only when the number of rounds $T$ is large enough.

While having a clear advantage in performance, our algorithms also run as fast as other online boosting algorithms. In fact, each iteration of all our algorithms except OSBoost.OCP can be carried out easily in $\mathcal{O}(N)$ time with $N$ weak learners. For OSBoost.OCP, an extra projection step is needed, for which we implement an $\mathcal{O}(N \log N)$-time algorithm by Duchi et al. (2008); in fact, a more sophisticated method in Duchi et al. (2008) can achieve this in expected $\mathcal{O}(N)$ time.

## 6. Conclusion

We propose a novel online boosting algorithm. The algorithm is simple in its formulation, but nevertheless carefully designed from the theoretical perspective to avoid many intrinsic difficulties when adapting boosting algorithms to the online setting. In particular, we define the notion of weak learning for online boosting, and exploit the notion to extend one promising offline boosting algorithm to its online version. We also tackle the problem of choosing a suitable number of weak learners with a careful use of established theoretical results.

---

[1] http://www.csie.ntu.edu.tw/~cjlin/libsvmtools/datasets.

Table 1. Comparison (error rate) with other online boosting algorithms using *Perceptron* weak leaner

| Data Set | # of examples | Perceptron | OzaBoost | OGBoost | OSBoost |
|---|---|---|---|---|---|
| Heart | 270 | 0.2489 | 0.2356 | 0.2267* | **0.2356** |
| Breast-Cancer | 683 | 0.0592 | 0.0501 | 0.0445* | **0.0466** |
| Australian | 690 | 0.2099 | 0.2012 | 0.1962 | **0.1872*** |
| Diabetes | 768 | 0.3216 | 0.3169* | 0.3313 | **0.3185** |
| German | 1000 | 0.3256 | 0.3364 | 0.3142* | **0.3148** |
| Splice | 3175 | 0.2717 | 0.2759 | 0.2625 | **0.2605*** |
| Mushrooms | 8124 | 0.0148 | 0.0080 | 0.0068 | **0.0060*** |
| Adult | 48842 | 0.2093 | 0.2045 | 0.2080 | **0.1994*** |
| Ijcnn1 | 141691 | 0.1070 | 0.1014 | 0.1046 | **0.0943*** |
| WebPage | 412943 | 0.0225 | 0.0203 | 0.0205 | **0.0182*** |
| Cod-RNA | 488565 | 0.2096 | 0.2170 | 0.2241 | **0.2075*** |
| Covertype | 581012 | 0.3437 | 0.3449 | 0.3482 | **0.3334*** |

Table 2. Comparison (error rate) with other online boosting algorithms using *Naive Bayes* weak leaner

| Data Set | Naive Bayes | OzaBoost | OGBoost | OSBoost |
|---|---|---|---|---|
| Heart | 0.1904 | 0.2570 | 0.3037 | 0.2059* |
| Breast-Cancer | 0.0474 | 0.0635 | 0.1004 | 0.0489* |
| Australian | 0.1751 | 0.2133 | 0.2826 | 0.1849* |
| Diabetes | 0.2664 | 0.3091 | 0.3292 | **0.2622*** |
| German | 0.2988 | 0.3206 | 0.3598 | **0.2730*** |
| Splice | 0.2520 | 0.1563 | 0.1863 | **0.1370*** |
| Mushrooms | 0.0076 | 0.0049 | 0.0229 | **0.0029*** |
| Adult | 0.2001 | 0.1912 | 0.1878 | **0.1581*** |
| Ijcnn1 | 0.1040 | 0.0805 | 0.0773 | **0.0764*** |
| Web Page | 0.0339 | 0.0221 | 0.0184* | **0.0189** |
| Cod-RNA | 0.2206 | 0.0796 | 0.0568* | **0.0581** |
| Covertype | 0.3518 | 0.3293* | 0.3732 | 0.3634 |

Table 3. Results (error rate) of our 3 variant algorithms (with *Perceptron*)

| Data Set | OSBoost | OSBoost.OCP | OSBoost.EXP |
|---|---|---|---|
| Heart | 0.2356 | **0.2311** | 0.2407 |
| Breast-Cancer | 0.0466 | 0.0515 | **0.0451** |
| Australian | 0.1872 | 0.2078 | **0.1852** |
| Diabetes | 0.3185 | 0.3315 | 0.3193 |
| German | 0.3148 | 0.3174 | **0.3090** |
| Splice | 0.2605 | **0.2590** | 0.2645 |
| Mushrooms | 0.0060 | 0.0062 | 0.0062 |
| Adult | 0.1994 | **0.1991** | **0.1991** |
| Ijcnn1 | 0.0943 | 0.0945 | 0.0949 |
| WebPage | 0.0182 | **0.0181** | 0.0182 |
| Cod-RNA | 0.2075 | **0.2059** | 0.2071 |
| Covertype | 0.3334 | 0.3338 | 0.3341 |

Table 4. Results (error rate) of our 3 variant algorithms (with *Naive Bayes*)

| Data Set | OSBoost | OSBoost.OCP | OSBoost.EXP |
|---|---|---|---|
| Heart | 0.2059 | 0.2852 | **0.2022** |
| Breast-Cancer | 0.0489 | 0.0665 | **0.0442** |
| Australian | 0.1849 | 0.2629 | **0.1838** |
| Diabetes | 0.2622 | 0.3284 | **0.2482** |
| German | 0.2730 | 0.3300 | 0.2796 |
| Splice | 0.1370 | 0.1615 | 0.1426 |
| Mushrooms | 0.0029 | 0.0045 | 0.0032 |
| Adult | 0.1581 | 0.1711 | 0.1582 |
| Ijcnn1 | 0.0764 | 0.0851 | 0.0770 |
| WebPage | 0.0189 | **0.0171** | 0.0189 |
| Cod-RNA | 0.0581 | **0.0484** | 0.0583 |
| Covertype | 0.3634 | 0.3646 | **0.3408** |

The proposed algorithm is not only solid in its theoretical justifications, but leads to promising experimental results. We demonstrate that the proposed algorithm can indeed boost online weak learners on real-world data sets. Most importantly, the proposed approach is significantly better than existing approaches while they do not come with such solid theoretical justifications. Future works include extending the proposed approach to other problems, including online multi-class or multi-label classification.

## Acknowledgments

We thank Ching-Hua Yu for providing the idea of using the framework of *Predicting with Expert Advice*. We also thank the anonymous reviewers for valuable suggestions. This work is supported by the National Science Council (NSC 100-2628-E-002-010 and NSC 100-2221-E-001-008-MY3) of Taiwan.